\documentclass{article}

\usepackage{microtype}
\usepackage{graphicx}
\usepackage{subfigure}
\usepackage{booktabs} % for professional tables

\usepackage{hyperref}

\usepackage[arxiv]{icml2025}

\usepackage{amsmath}
\usepackage{amssymb}
\usepackage{mathtools}
\usepackage{amsthm}

\usepackage[capitalize,noabbrev]{cleveref}

\theoremstyle{plain}

\theoremstyle{definition}

\theoremstyle{remark}

\DeclareMathOperator{\mhsa}{MHSA}
\DeclareMathOperator{\mhca}{MHCA}

\usepackage[textsize=tiny]{todonotes}

\PassOptionsToPackage{table,dvipsnames}{xcolor}

\usepackage{times}
\usepackage{epsfig}  % Remove duplicate
\usepackage{graphicx} % Remove duplicate
\usepackage{amsmath}
\usepackage{amssymb}
\usepackage{todonotes}
\usepackage{lineno}
\usepackage{colortbl}
\usepackage{caption}
\usepackage{subcaption}
\usepackage{booktabs}
\usepackage{xcolor}  % Load xcolor once after passing options
\usepackage{soul}
\usepackage{xr}
\usepackage{multirow}
\usepackage{algorithm}
\usepackage{makecell}
\usepackage[accsupp]{axessibility}

\definecolor{darkgreen}{rgb}{0.0, 0.4, 0.0}
\definecolor{orange}{rgb}{1.0, 0.49, 0.0}
\definecolor{purple}{rgb}{0.54, 0.17, 0.89}
\definecolor{ForestGreen}{RGB}{34,139,34}
\definecolor{ModernBlue}{RGB}{0,153,153}
\definecolor{LightModernBlue}{RGB}{0,204,204}
\definecolor{DarkPink}{RGB}{204,0,102}
\definecolor{Gray}{gray}{0.9}
\definecolor{LightCyan}{rgb}{0.88,1,1}

\newcommand{\modulename}{CASIM}

\icmltitlerunning{{\modulename}: Composite Aware Semantic Injection for Text to Motion Generation}
\begin{document}

\twocolumn[
%%%% Some candidates for title, should tailor to the taste of ICML reviewers
% \icmltitle{{\modulename}: Rethinking the Semantic Injection for Text to Motion Generation}
\icmltitle{{\modulename}: Composite Aware Semantic Injection for Text to \\ Motion Generation}
% \icmltitle{Rethinking the Injection of CLIP Semantics for Text to Motion Generation}
% \icmltitle{MODELNAME: An Attention-based Semantic Injection for Text to Motion Generation}



\icmlsetsymbol{equal}{*}

\begin{icmlauthorlist}
\icmlauthor{Che-Jui Chang}{equal,ru,amzn}
\icmlauthor{Qingze (Tony) Liu}{equal,ru}
\icmlauthor{Honglu Zhou}{crm}
\icmlauthor{Vladimir Pavlovic}{ru}
\icmlauthor{Mubbasir Kapadia}{rblx}
\end{icmlauthorlist}

\icmlaffiliation{ru}{Rutgers University}
\icmlaffiliation{amzn}{Amazon}
\icmlaffiliation{crm}{Salesforce AI Research}
\icmlaffiliation{rblx}{Roblox}

\icmlcorrespondingauthor{Qingze (Tony) Liu}{tony.liu@rutgers.edu}
\icmlcorrespondingauthor{Che-Jui Chang}{chejui.chang@rutgers.edu}

\icmlkeywords{Human Motion Generation, Text to Motion Generation, Semantic Injection, Composite Aware Semantic Injection, T2M, {\modulename}}

\vskip 0.3in
]



\printAffiliationsAndNotice{\icmlEqualContribution} % otherwise use the standard text.

\begin{abstract}
% \vspace{-2pt}
Recent advances in generative modeling and tokenization have driven significant progress in text-to-motion generation, leading to enhanced quality and realism in generated motions. However, effectively leveraging textual information for conditional motion generation remains an open challenge. We observe that current approaches, primarily relying on fixed-length text embeddings (e.g., CLIP) for global semantic injection, struggle to capture the composite nature of human motion, resulting in suboptimal motion quality and controllability. To address this limitation, we propose the Composite Aware Semantic Injection Mechanism ({\modulename}), comprising a composite aware text encoder and a text-motion aligner that learns the dynamic correspondence between text and motion tokens. 
Notably, {\modulename} is model and representation-agnostic, readily integrating with both autoregressive and diffusion-based methods. Experiments on HumanML3D and KIT benchmarks demonstrate that {\modulename} consistently improves motion quality, text-motion alignment, and retrieval scores across state-of-the-art methods. Qualitative analyses further highlight the superiority of our composite aware approach over fixed-length semantic injection, enabling precise motion control from text prompts and stronger generalization to unseen text inputs.
Our code is available at our
project page: \url{https://cjerry1243.github.io/casim_t2m}.

\end{abstract}

\vspace{-10pt}
\section{Introduction} \label{sec:intro}
\vspace{-3pt}

\begin{figure}[!ht]
    \centering
    \includegraphics[width=\linewidth]{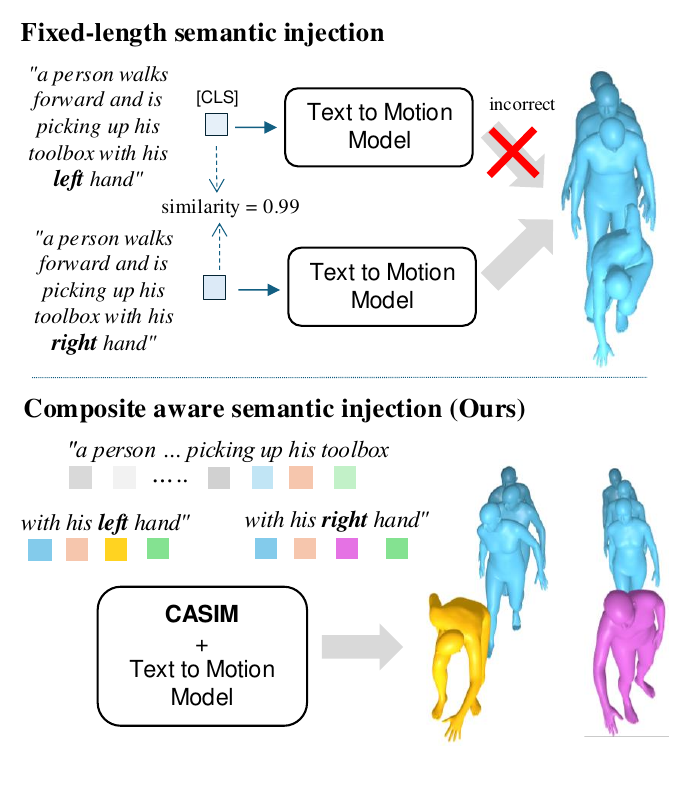}
    \vspace{-35pt}
    \caption{(\textit{\textbf{Top}}) Fixed-length semantic injection, which primarily relied on the [CLS] token embedding from CLIP~\cite{radford2021learning} to represent the entire text prompt, fails to capture the subtle differences in individual words. As a result, it generates highly similar motions from distinct text prompts. (\textit{\textbf{Bottom}}) Our Composite aware semantic injection method allows each motion frame to dynamically attend to every word token (e.g., ``left'' or ``right'' hand), enhancing the motion-text correspondence.}
    \vspace{-10pt}
\label{fig:motivation}
\end{figure}

Human motion generation from text descriptions has gained increasing attention from the community, due to its immense potential for animating and editing lifelike motions with free-form text prompts \cite{MotionFix, goel2024iterative}.
The advancements in generative modeling techniques and efficient motion tokenization methods have been driving the progress in text to motion generation, with particular focus on enhancing motion quality.
Several diffusion based \cite{tevet2023human, zhang2022motiondiffuse, guo2023momask, chen2023executing, CondMDICohan2024, chang2024learning} and autoregressive based \cite{zhang2023generating, huang2024como, jiang2024motiongpt, chen2024motionllm} methods have shown impressive capability in generating realistic and diverse motions from the input prompts. 
Typically, these methods leverage the pre-trained CLIP model \cite{radford2021learning} to summarize the whole motion description with various lengths and complexity into a fixed-length embedding, denoted as [CLS]. 
The compressed text embedding is then used as a conditional vector, injected into the motion generation model.
% The CLIP text encoder uses a fixed length high-dimensional token [CLS] to summarize motion descriptions with various lengths and complexity.
The assumption is that the [CLS] token contains rich and global semantics that are suitable for motion generation. 
While this approach may seem to work to a certain degree for most existing models, the composite nature of human motions, causal textual order, and the fine-grained alignment of the text and motion tokens are poorly preserved. 
As illustrated in Fig.~\ref{fig:motivation}, the fixed-length semantic injection struggles to produce distinguishable conditional vectors for longer descriptions, leading to poor generalization and limited control over generated motions.

To address these limitations, we introduce {\modulename}, Composite Aware Semantic Injection Mechanism (Sec.~\ref{sec:casim}). 
It comprises a composite aware text encoder and a text-motion aligner that learns dynamic alignment between each text tokens and motion frames.
Compared with fixed-length approaches, {\modulename} offers a more generalized and robust architecture, allowing each textual component to influence either specific composites or global attributes of the motion sequence.
Another key advantage of {\modulename} is its model-agnostic nature -- it can be easily integrated with both diffusion-based and autoregressive-based approaches while consistently improving text-motion matching and retrieval accuracy.
Furthermore, our method is compatible with various motion representations, from raw motion~\cite{tevet2023human} to different quantization approaches including VQVAE~\cite{zhang2023generating}, PoseCode~\cite{huang2024como}, and RVQ-VAE~\cite{guo2023momask}.

% What did we achieve?
Our experimental results (Sec.~\ref{sec:experiments}) demonstrate remarkable quantitative improvements in FID, R-precision, and Multimodality Distance, on HumanML3D and KIT benchmarks, as we apply our {\modulename} to 5 SOTA methods, including MDM~\cite{tevet2023human}, MotionDiffuse~\cite{zhang2022motiondiffuse}, T2MGPT~\cite{zhang2023generating}, COMO~\cite{huang2024como}, and MoMask~\cite{guo2023momask}.
{\modulename} can also improve motion quality and text-motion alignment for long-term human motion generation \cite{shafir2024human, lee2024t2lm}.
Analysis of attention weights inside {\modulename} verifies that our semantic injection mechanism effectively learns the expected dynamic matching between text tokens and motion frames, validating our design principles.
Qualitative results further demonstrate the superiority of {\modulename} over fixed-length semantic injection methods, showing precise motion control from text prompts and robust generalization to unseen text inputs.

\begin{figure*}[!t]
    \centering
    \includegraphics[width=\textwidth]{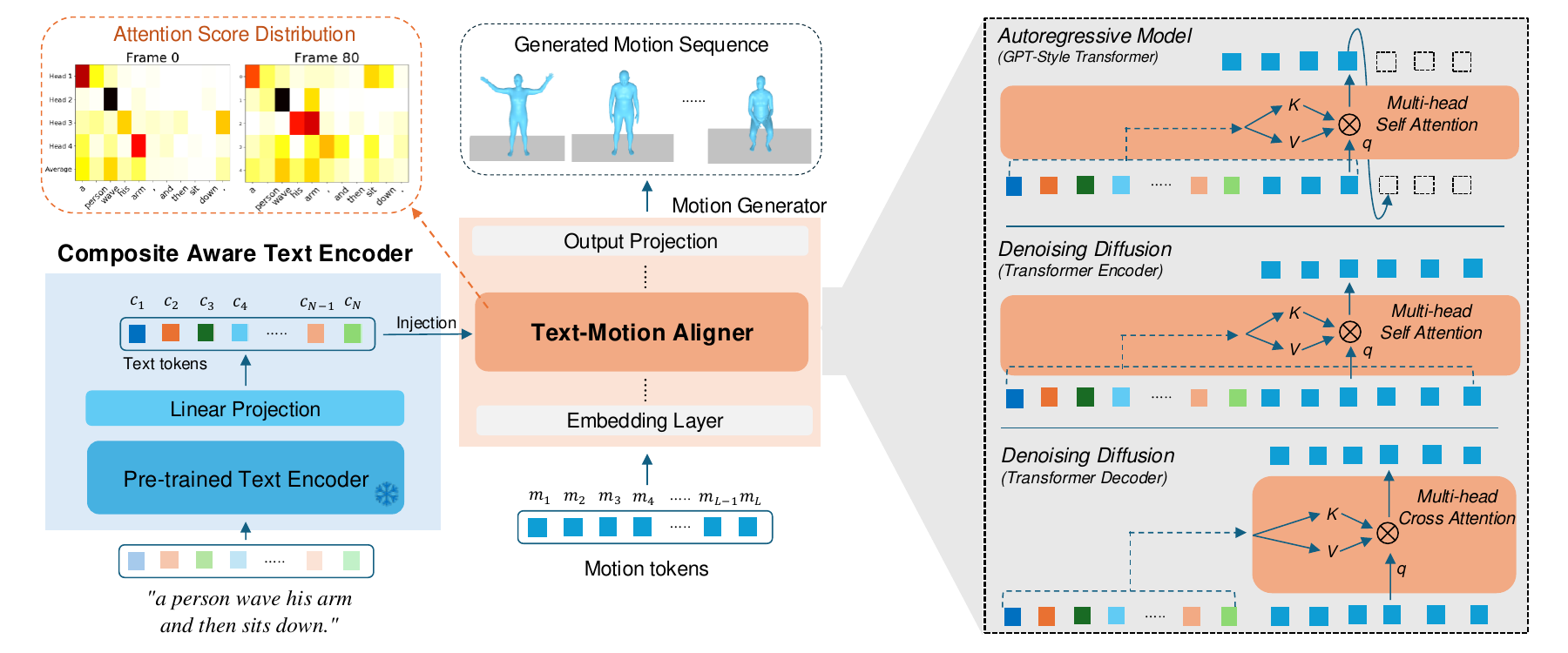}
    \vspace{-15pt}
    \caption{{\modulename} consists of two major components: Composite Aware Text Encoder (Left) for extracting granular word-level embeddings and Text-Motion Aligner (Middle) for aligning motion embeddings with relevant textual embeddings inside a motion generator. The attention score distribution between different motion tokens and the text tokens is visualized on the upper left. 
    % Red box show case a example of attention allocation between a subset of word and motion tokens.
    The Text-Motion Aligner can be integrated with three genres of motion generation models (Right).}
    \vspace{-5pt}
\label{fig:CASIM}
\end{figure*}

\vspace{-5pt}
\section{Related Works} \label{sec:related_works}
\vspace{-3pt}

\subsection{Human Motion Generation}
\vspace{-3pt}
% What are various tasks of X to motion generation
Generating realistic human motions has been a longstanding challenge in computer graphics and computer vision.
The field has evolved to embrace various input modalities and conditions for motion synthesis.
Image and video-based approaches have focused primarily on human pose and shape estimation~\cite{zhao2019semantic} and 3D body tracking~\cite{stathopoulos2024score}, enabling motion reconstruction and prediction from visual inputs.
Audio-driven motion generation is another important direction, with music-to-dance synthesis~\cite{alexanderson2023listen} and speech-to-gesture generation~\cite{chang2022ivi, chang2023importance} 
showing promising results in creating natural human movements that align with acoustic signals.
Text-to-motion generation has gained significant attention, as it offers intuitive control over motion synthesis through free-form text prompts~\cite{Guo_2022_CVPR}.
Scene-aware motion generation considers environmental constraints and spatial relationships, enabling the synthesis of contextually appropriate movements within 3D environments~\cite{cen2024text_scene_motion}.
Generating a coordinated group of human motions and interactions~\cite{chang2024learning, chang2024equivalency} 
has recently emerged as a novel research directionh adds another layer of difficulty to single-person motion generation due to the complex human interactions.
Lastly, several works \cite{li2024unimotion, zhou2024avatargpt} have attempted to unify motion generation, planning, and understanding in a single framework, extending the capabilities of large language models to human motion domains.

% \vspace{-5pt}
\subsection{Text-to-Motion Generation Models}
% \vspace{-3pt}

Text-to-motion generation models can be broadly categorized into two approaches: diffusion-based and autoregressive-based methods. 
Diffusion-based models leverage an iterative denoising scheme \cite{dhariwal2021diffusion} to generate motions from textual conditions. 
Notable works include MDM \cite{tevet2023human}, MotionDiffuse \cite{zhang2022motiondiffuse}, MLD \cite{chen2023executing}, GMD \cite{karunratanakul2023gmd}, FineMoGen \cite{zhang2023finemogen}, and GraphMotion \cite{jin2023act}. For example, MDM employs a transformer encoder within each diffusion step, processing concatenated motion frames with text and timestamp embeddings. While MLD adopts a similar architecture, it operates in a learned latent space by compressing motion sequences into fixed-length representations.
Autoregressive-based approaches, including T2M \cite{zhang2023generating}, TM2T \cite{guo2022tm2t}, MotionGPT \cite{jiang2024motiongpt}, MotionLLM \cite{chen2024motionllm}, T2MGPT \cite{zhang2023generating}, and CoMo \cite{huang2024como}, generate motions sequentially and typically require effective motion tokenization strategies. 
For instance, T2MGPT utilizes VQVAE \cite{van2017neural} for motion tokenization and implements a decoder-only architecture for motion token generation. CoMo follows a similar generator architecture but distinguishes itself by adopting heuristics-based posecodes \cite{delmas2022posescript} as its discrete motion representation.

\vspace{-2mm}
\subsection{Semantic Injection for Motion Generation}
\vspace{-1mm}

% More deeper dive into semantic injection and alignment for specific models, 
% Start with clip-based fixed length embedding
% such as finemogen, augmented prompts from chatgpt, CoMo keywords, heuristic-based hierarchical GraphMotion semantic injection 
% and how we differs from theirs.
For text-to-motion generation, the input prompts are typically encoded into a latent space with a well-trained text encoder before being passed to motion generation models. 
Previous works, such as MDM, MLD, T2MGPT and CoMo, leverage the pretrained CLIP text embedding to represent the full text prompt.
% as one of the inputs to their transformer-based encoder or decoder.
CoMo \cite{huang2024como} and FGMDM \cite{shi2023generating} includes several fine-grained keywords and descriptions from GPT4 \cite{openai2024gpt4technicalreport} as augmented prompts for motion generation.
Finemogen \cite{zhang2023finemogen} targets at fine-grained motion control and editing, by specifying the spatial and temporal motion descriptions.
GraphMotion \cite{jin2023act} parses the sentence structure into a hierarchical semantic graph for any given input texts. 
It utilizes a graph reasoning network and a coarse-to-fine diffusion model for motion generation.
Our {\modulename} is conceptually similar to GraphMotion as both are aimed at strengthening the text-motion correspondence by design. 
However, GraphMotion uses heuristic knowledge to create a static semantic graph and is only tied to its coarse-to-fine model.
{\modulename} learns the dynamic alignment and hierarchical structure in a soft manner. 
It can also be flexibly integrated with the most existing models.

\vspace{-5pt}
\section{Composite Aware Semantic Injection} \label{sec:casim}
\vspace{-1pt}

{\modulename}, Composite Aware Semantic Injection Mechanism, is designed to capture fine-grained semantic relationships between text descriptions and motion sequences. It preserves the composite nature of human motions and their causal textual ordering and allows each motion frame to dynamically align with relevant textual components at different granularities.
{\modulename} consists of two principal components: a composite aware text encoder and a text-motion aligner. 
{\modulename} exhibits model-agnostic properties, as it is applicable to both autoregressive and diffusion-based motion generators, which represent the two predominant genres for state-of-the-art models. 
In Section \ref{sub:formulation}, we introduce the formulation of {\modulename}, detailing its two major components. 
Section \ref{sub:autoregressive} describes for autoregressive motion generators and how {\modulename} is integrated. 
Section \ref{sub:duffsion} discusses diffusion-based motion generators and explains how to adopt {\modulename} 
% is adapted
in this framework.

% Key cross attention module + clip-based token embedding as a semantic injection module
\vspace{-3pt}
\subsection{{\modulename} Formulation} % module name
\label{sub:formulation}
\vspace{-3pt}

\textbf{Composite Aware Text Encoder.} Unlike traditional approaches that compress text descriptions into fixed-length [CLS] token, our text encoder preserves composite aware semantics through individual token-level embeddings. 
As shown in Fig. \ref{fig:CASIM} (left), the encoder comprises a pretrained text encoder inside which are blocks of multihead self-attention layers that learn the latent features for the input text. 
% , crucial for text-driven motion generation. 
We leverage the pre-trained text encoder from CLIP~\cite{radford2021learning}, project the latent encoder output to another embedding space, and inject the resulting token embeddings to the motion generator.
% The resulting text embeddings serve as conditions for the text-motion aligner. 
% The features capture both sentence-level context and word-level details.
Compared to fixed-length CLIP embeddings, our injection method preserve the semantics as granular as the token-level, which are essential for composite aware motion generation.

\textbf{Text-Motion Aligner.} The text-motion aligner is the core design of {\modulename} and can be integrated inside various motion generation models.
Specifically, it establishes dynamic correspondence between motion frames and text tokens using multi-head attention (MHA). 
As illustrated in Fig.~\ref{fig:CASIM} (middle), each motion token is used as query to attend with the keys and values obtained from all the text tokens. 
The subsequent motion embeddings then are updated through the attention-weighted aggregation from all relevant text tokens. 
Depending on the motion generation approach, the aligner employs multihead self-attention (MHSA) for autoregressive generation or multihead cross-attention (MHCA) for diffusion-based generation. 
\vspace{-3pt}
\subsection{{\modulename} for Autoregressive Motion Generation}
\label{sub:autoregressive}
\vspace{-1pt}

Autoregressive motion generation approaches, inspired by language modeling principles \cite{jiang2024motiongpt, chen2024motionllm}, typically follow a two-stage process. First, they learn a discrete representation of motions through tokenization. Given a motion sequence $X=\{x_1,\dots,x_T\}$, an encoder-decoder network is trained to transform it into a sequence of motion tokens $M=\{m_1,\dots,m_{T/l}\}$, where each subsequence of $l$ frames $X_l=\{x_i,\dots,x_{i+l}\}$ is mapped to a discrete token $m_i = \mathcal{E}(X_l)$. The decoder $\mathcal{D}$ learns to reconstruct the original motion: $\hat{X}=\mathcal{D}(M)$.
In the second stage, these methods train an autoregressive model to predict motion tokens sequentially conditioned on a text prompt $C$. The generation process is as follows:
\begin{equation}
P(M|C) = P(m_1|C)\prod_{i=1}^{T/l}P(m_{i+1}|m_1,\dots,m_i,C)
\end{equation}
where each new motion token is predicted based on both the text condition and previously generated tokens. The final motion is obtained by passing the generated tokens through the pretrained decoder $\mathcal{D}$.

% For autoregressive generation, {\modulename} utilizes MHSA as the text-motion aligner at each generation step $P(m_i|m_j^{<i},C)=MHSA(M^{<i}\oplus C)$. As shown in Figure \ref{fig:CASIM} (right), we concatenate the text token sequence $C={c_1,\dots,c_L}$ with the previously generated motion tokens $M^{<i}$. This concatenated sequence is processed by an encoder-only transformer with stacked MHSA blocks, which performs self-attention between motion and text tokens to guide the generation of the next motion token.

For autoregressive generation, {\modulename} leverages a GPT-style transformer \cite{radford2018improving, zhang2023generating} with MHSA blocks for the text-motion aligner to predict the motion tokens.
At generation step $i$, we concatenate the text token sequence $C=\{c_1,\dots,c_N\}$ with previously generated motion tokens $M^{<i}$, denoted as $C\oplus M^{<i}$. 
As shown in Figure \ref{fig:CASIM} (right), this concatenated sequence is processed by an GPT-style transformer with stacked MHSA blocks that enable dynamic interaction between motion and text tokens. 
The probability of generating the next token is computed as:
\begin{equation}
P(m_i|m_j^{<i},C)=\sigma(\mhsa(C \oplus M^{<i}))),
\end{equation} where $\sigma$ represents the softmax function.

This generation step is performed iteratively until the end-of-sequence token is predicted.
The integration of {\modulename} with autoregressive models enables each motion token to be generated with awareness of both previously generated motions and the full text description.

\begin{table}[t!]
% \small
% \footnotesize
% \scriptsize
\fontsize{7.5pt}{7.5pt}\selectfont
  \aboverulesep=0ex
  \belowrulesep=0.5ex 
\setlength{\tabcolsep}{4.2pt}
\centering
\caption{Results on the HumanML3D dataset. The original and {\modulename}-integrated models are shaded.}
\vspace{-5pt}
\begin{tabular}{lcccccc}
\toprule
\multirow{2}{*}{Method}  & \multicolumn{3}{c}{R-Precision$\uparrow$} & \multirow{2}{*}{FID$\downarrow$} & \multirow{2}{*}{MM-Dist.$\downarrow$} & \multirow{2}{*}{Div.$\uparrow$} \\
 \cmidrule{2-4}
 & Top1 & Top2 & Top3 &  &  &  \\
\midrule 
T2M & 0.457 & 0.639 & 0.740 & 1.067 & 3.340 & 9.188 \\
TM2T & 0.424 & 0.618 & 0.729 & 1.501 & 3.467 & 8.589 \\
MotionDiffuse & 0.491 & 0.681 & 0.782 & 0.630 & 3.113 & 9.410 \\
MLD & 0.469 & 0.659 & 0.760 & 0.532 & 3.282 & 9.570 \\
\rowcolor{Gray} MDM & 0.455 & 0.645 & 0.749 & 0.489 & 3.330 & 9.920 \\
\rowcolor{Gray} T2MGPT & 0.491 & 0.680 & 0.775 & 0.116 & 3.118 & 9.761 \\
FineMoGen & 0.504 & 0.690 & 0.784 & 0.151 & 2.998 & 9.263 \\
MotionGPT & 0.492 & 0.681 & 0.778 & 0.232 & 3.096 & 9.528 \\
CoMo & 0.502 & 0.692 & 0.790 & 0.262 & 3.032 & \textbf{9.936} \\
GraphMotion & 0.504 & 0.699 & 0.785 & 0.116 & 3.070 & 9.692 \\
\midrule
\rowcolor{Gray}
\modulename-MDM & 0.502 & 0.694 & 0.793 & 0.165 & 3.020 & 9.394 \\
\rowcolor{Gray}
\modulename-T2MGPT & \textbf{0.539} & \textbf{0.730} & \textbf{0.823} & \textbf{0.105} & \textbf{2.838} & 9.785 \\
%  & CoMo  & $\checkmark$ & 0.545 & 0.741 & 0.835 & 0.200 & 2.747 & 9.887 \\
\bottomrule
\label{tab:quant_res1}
\end{tabular}
\vspace{-5pt}
\end{table}

\begin{table}[t!]
% \small
% \footnotesize
% \scriptsize
\fontsize{7.5pt}{7.5pt}\selectfont
  \aboverulesep=0ex
  \belowrulesep=0.5ex 
\setlength{\tabcolsep}{4.2pt}
\centering
% \caption{Evaluation of generated motion on HumanML3D and KIT-ML dataset. {\modulename} integrated models is shaded}
\caption{Results on the KIT-ML dataset.
The original and {\modulename}-integrated models are shaded.}
\vspace{-5pt}
\begin{tabular}{lcccccc}
\toprule
\multirow{2}{*}{Method}  & \multicolumn{3}{c}{R-Precision$\uparrow$} & \multirow{2}{*}{FID$\downarrow$} & \multirow{2}{*}{MM-Dist.$\downarrow$} & \multirow{2}{*}{Div.$\uparrow$} \\
 \cmidrule{2-4}
 & Top1 & Top2 & Top3 &  &  &  \\
\midrule 
% Real & 0.424 & 0.649 & 0.779 & 0.031 & 2.788 & 11.080 \\
T2M & 0.370 & 0.569 & 0.693 & 2.770 & 3.401 & 10.910 \\
TM2T & 0.280 & 0.463 & 0.587 & 3.599 & 4.591 & 9.473 \\
MotionDiffuse & 0.417 & 0.621 & 0.739 & 1.954 & 2.958 & 11.100 \\
MLD & 0.390 & 0.609 & 0.734 & 0.404 & 3.204 & 10.800 \\
\rowcolor{Gray} MDM & 0.164 & 0.291 & 0.396 & 0.497 & 9.191 & 10.847 \\
\rowcolor{Gray} T2MGPT & 0.402 & 0.619 & 0.737 & 0.717 & 3.053 & 10.862 \\
FineMoGen & 0.432 & 0.649 & 0.772 & \textbf{0.178} & 2.869 & 10.850 \\
MotionGPT & 0.366 & 0.558 & 0.680 & 0.510 & 3.527 & 10.350 \\
CoMo & 0.422 & 0.638 & 0.765 & 0.332 & 2.873 & 10.950 \\
GraphMotion & 0.429 & 0.648 & 0.769 & 0.313 & 3.076 & 11.120 \\
\midrule
\rowcolor{Gray}
\modulename-MDM & \textbf{0.448} & \textbf{0.665} & \textbf{0.784} & 0.354 & \textbf{2.684} & \textbf{11.179} \\
\rowcolor{Gray}
\modulename-T2MGPT & 0.412 & 0.627 & 0.751 & 0.577 & 2.986 & 10.829 \\
\bottomrule
\label{tab:quant_res2}
\end{tabular}
\vspace{-5pt}
\end{table}

\vspace{-3pt}
\subsection{{\modulename} for Motion Diffusion Generation}
\label{sub:duffsion}
\vspace{-1pt}

Unlike autoregressive approaches, motion diffusion models generate motions through iterative denoising of a Gaussian noise sequence $X^\tau$. Let $T$ denote the sequence length and $D$ denote the motion dimension, then $X^\tau \in \mathbb{R}^{T \times D}$ represents the noised motion at diffusion step $\tau$. Given a noise-free motion sequence $X^0$, the diffusion process at each step $t$ samples the denoised motion according to:
\begin{equation}
X^{\tau-1} \sim \mathcal{N}(\mu_{\theta}(X^\tau, \tau, C), \Sigma(\tau)),
\end{equation}
where $\Sigma(\tau)$ is a fixed variance schedule, $C \in \mathbb{R}^{L \times d}$ denotes the text tokens with sequence length $L$ and embedding dimension $d$. The mean term $\mu_{\theta}(X^\tau, \tau, C)$ can be parameterized as $\mu_{\theta}(X^\tau, \hat{X^0})$, where $\hat{X^0}=G_\theta(X^\tau, \tau, C)$ is the predicted noise-free motion by the denoiser $G_\theta$.

While there are no constraints on how the denoiser should be designed as long as the input and output shape matches, the transformer encoder and decoder are the two widely adopted options in the literature.
To integrate {\modulename} with these denoiser variants in diffusion models, we adapt our text-motion aligner to employ MHSA for the transformer encoder and MHCA for the transformer decoder.

\textbf{Denoising with Transformer Encoder.} In the encoder-based approach, we first augment the text embeddings $C$ with diffusion step embeddings $TE(\tau)$, where $TE(\cdot)$ denotes the positional encoding function. The augmented text embeddings are then concatenated with the current motion embeddings $X^t$, and processed through MHSA blocks:
\begin{equation}
\hat{X^0} = \mhsa( (C+TE(\tau))\oplus X^\tau),
\end{equation}
where $\oplus$ denotes sequence concatenation.

\textbf{Denoising with Transformer Decoder.} The decoder-based variant first processes motion features through self-attention layers, enabling each motion frame to attend to other frames. It then applies cross-attention (MHCA) between the processed motion embeddings $X^\tau$ and the timestep-augmented text embeddings:
\begin{equation}
\hat{X^0}  = \mhca(X^\tau, C+TE(\tau)),
\end{equation}

Both formulations enable dynamic text-motion alignment during the denoising process. 
We analyze the performance of all model genres and variants in Section \ref{sub:quantitative}.

\begin{table}[t!]
\fontsize{7.5pt}{7.5pt}\selectfont
  \aboverulesep=0ex
  \belowrulesep=0.5ex 
\setlength{\tabcolsep}{4.2pt}
\centering
\caption{Quantitative results for various text-to-motion methods with {\modulename} on HumanML3D dataset.}
\vspace{-5pt}
\begin{tabular}{lcccccc}
\toprule
\multirow{2}{*}{Method} & \multirow{2}{*}{\modulename} & \multicolumn{3}{c}{R-Precision$\uparrow$} & \multirow{2}{*}{FID$\downarrow$} & \multirow{2}{*}{MM-Dist.$\downarrow$} \\
 \cmidrule{3-5}
 & & Top1 & Top2 & Top3 &  &   \\
\midrule 
\multirow{2}{*}{MotionDiffuse}  & & 0.491 & 0.681 & 0.782 & 0.630 & 3.113 \\
 & $\checkmark$  & \textbf{0.506} & \textbf{0.700} & \textbf{0.799} & \textbf{0.565} & \textbf{2.981} \\
\midrule
\multirow{2}{*}{MDM}  & & 0.471 & 0.661 & 0.760 & 0.325 & 3.249 \\
  & $\checkmark$ & \textbf{0.502} & \textbf{0.694} & \textbf{0.793} & \textbf{0.165} & \textbf{3.020} \\
\midrule
\multirow{2}{*}{T2MGPT}  & & 0.484 & 0.672 & 0.770 & 0.117 & 3.153 \\
  & $\checkmark$ & \textbf{0.539} & \textbf{0.730} & \textbf{0.823} & \textbf{0.105} & \textbf{2.838} \\
\midrule
\multirow{2}{*}{CoMo}  & & 0.502 & 0.692 & 0.790 & 0.262 & 3.032 \\
  & $\checkmark$ & \textbf{0.545} & \textbf{0.741} & \textbf{0.835} & \textbf{0.200} & \textbf{2.747} \\
\midrule
 % & MoMask  & & 0.521 & 0.713 & 0.807 & 0.045 & 2.958 & 9.645 \\
\multirow{2}{*}{MoMask}  & & 0.510 & 0.703 & 0.801 & 0.064 & 2.997 \\
  & $\checkmark$ & \textbf{0.532} & \textbf{0.719} & \textbf{0.813} & \textbf{0.057} & \textbf{2.911} \\
% \midrule
\bottomrule
\vspace{-5pt}

\label{tab:quant_res3}
\end{tabular}
\end{table}

\begin{table}[t!]

\fontsize{7.5pt}{7.5pt}\selectfont
  \aboverulesep=0ex
  \belowrulesep=0.5ex 
\setlength{\tabcolsep}{5.5pt}
\centering
\caption{Quantitative results for various text-to-motion methods with {\modulename} on KIT-ML dataset.}
\vspace{-7.5pt}
\begin{tabular}{lcccccc}
\toprule
\multirow{2}{*}{Method} & \multirow{2}{*}{\modulename} & \multicolumn{3}{c}{R-Precision$\uparrow$} & \multirow{2}{*}{FID$\downarrow$} & \multirow{2}{*}{MM-Dist.$\downarrow$} \\
 \cmidrule{3-5}
 & & Top1 & Top2 & Top3 &  &   \\
\midrule 

\multirow{2}{*}{MDM} &  & 0.164 & 0.291 & 0.396 & 0.497 & 9.191 \\
 & $\checkmark$ & \textbf{0.448} & \textbf{0.665} & \textbf{0.784} & \textbf{0.354} & \textbf{2.684} \\
\midrule 
\multirow{2}{*}{T2MGPT} & & 0.402 & 0.619 & 0.737 & 0.717 & 3.053 \\
 & $\checkmark$ & \textbf{0.412} & \textbf{0.627} & \textbf{0.751} & \textbf{0.577} & \textbf{2.986} \\
\midrule
\multirow{2}{*}{CoMo} &  & 0.399 & - & - & \textbf{0.399} & 2.898 \\
 & $\checkmark$ & \textbf{0.422} & \textbf{0.641} & \textbf{0.762} & 0.408 & \textbf{2.852} \\
\bottomrule
\label{tab:quant_res4}
\vspace{-5pt}
\end{tabular}
\end{table}

\vspace{-5pt}
\section{Experiments} \label{sec:experiments}
\vspace{-3pt}

We evaluate {\modulename} on two standard datasets through extensive quantitative and qualitative analyses. Section \ref{sub:eval_setup} describes our experimental setup, followed by quantitative results in Section \ref{sub:quantitative}, qualitative analysis in Section \ref{sub:qualitative}, and extension to long-form generation in Section \ref{sub:long-term}.

\vspace{-3pt}
\subsection{Evaluation Setup}
\label{sub:eval_setup}
\vspace{-3pt}

\textbf{Datasets and Metrics.} 
We evaluate on HumanML3D \cite{Guo_2022_CVPR} and KIT Motion Language (KIT-ML) \cite{Plappert2016}.
HumanML3D is a large-scale motion language dataset, containing a total of 14,616 motions from AMASS \cite{mahmood2019amass} and HumanAct12 \cite{guo2020action2motion} datasets. 
Each motion is paired with 3 textual annotations, totaling 44,970
descriptions.
KIT-ML dataset consists of a total of 3,911 motions and 6,278 text annotations, providing a small-scale evaluation benchmark. 
The datasets are split into train-valid-test sets with a ratio of 0.8:0.05:0.15. 
Performance is measured using the following metrics: 
(a) \textit{Frechet Inception Distance (FID)}, which evaluates the overall motion quality by computing the distributional difference between the latent features of the generated motions and those of real motions from test set; 
(b) \textit{R-Precision}, which reports the retrieval accuracy between input text and generated motions from their latent space; 
(c) \textit{MM-Distance}, which reports the distance between the input text and generated motions at the latent space; and
(d) \textit{Diversity}, which assesses the diversity of all generated motions.
All metrics are computed using a separate text-motion matching network from \cite{Guo_2022_CVPR}.

\textbf{Baseline Models.} 
We conduct experiments for {\modulename} with five state-of-the-art (SOTA) models: 
T2MGPT \cite{zhang2023generating}, CoMo \cite{huang2024como}, MoMask \cite{guo2023momask},
MotionDiffuse \cite{zhang2022motiondiffuse}, 
and MDM \cite{tevet2023human}, covering various genres of motion generation models as well as both continuous and discrete motion representations.

T2MGPT and CoMo are \textbf{autoregressive} motion generation models, which first tokenize the motion sequence using a quantization-based encoder-decoder structure. 
% After the motions are processed into a sequence of tokens, they employ an autoregressive generator that takes the injected text semantics as the initial token.
We use the autoregressive form of {\modulename}, as detailed in Section \ref{sub:autoregressive}, in place of the fixed-length text injection for their motion token generation.
% Among them, T2MGPT and CoMo are two \textbf{autoregressive} motion generation models. Here, we adopt the autoregressive form of {\modulename} detailed in section \ref{sub:autoregressive} to replace the original autoregressive generator that takes the fix-length [CLS] token as text injection.
% MotionDiffuse and MDM are \textbf{diffusion-based} motion generation models, which undertake conditional denoising at each step of the diffusion process on the entire motion sequence. 
% The core architecture under each diffusion scheme is either a transformer encoder or decoder, that maps the noised motion into the denoised one of the same shape. 
% We apply the composite aware text injection for both the encoder and decoder variants of MDM, while for MotionDiffuse, we use the encoder-based semantic injection. 
MotionDiffuse and MDM are \textbf{diffusion-based} motion generation models that conditionally denoise full motion sequences at each diffusion step with the fixed-length semantic embedding. 
We apply the composite aware text injection for both the encoder and decoder variants of MDM, while for MotionDiffuse, we use the encoder-based semantic injection. 
MoMask employs a hierarchical motion quantization scheme and a multi-layer masked motion generation framework.  
We apply the encoder-based semantic injection for their M- and R- Transformers in our study.

All the other settings follow the baseline methods and hyperparameters remain unchanged.

\begin{table}[t!]
\fontsize{7.25pt}{7.25pt}\selectfont
  \aboverulesep=0ex
  \belowrulesep=0.5ex 
\setlength{\tabcolsep}{3.0pt}
\centering
\caption{Additional quantiative results on HumanML3D with varying architectures and configuration settings.}
\vspace{-5pt}
\begin{tabular}{llcccccc}
\toprule
\multirow{2}{*}{Method} & \multirow{2}{*}{Arch} & \multirow{2}{*}{\modulename} & \multicolumn{3}{c}{R-Precision$\uparrow$} & \multirow{2}{*}{FID$\downarrow$} & \multirow{2}{*}{MM-Dist.$\downarrow$} \\
 \cmidrule{4-6}
 & & &  Top1 & Top2 & Top3 &  &   \\
\midrule 
% MDM (Paper) & Enc & & 0.455 & 0.645 & 0.749 & 0.489 & 3.330 \\
\textcolor{gray}{MDM (Paper)}  & \textcolor{gray}{Enc} & & \textcolor{gray}{0.418} & \textcolor{gray}{0.604} & \textcolor{gray}{0.707} & \textcolor{gray}{0.489} & \textcolor{gray}{3.630} \\
\textcolor{gray}{MDM (Paper)} & \textcolor{gray}{Dec} & & \textcolor{gray}{--} & \textcolor{gray}{--} & \textcolor{gray}{0.608} & \textcolor{gray}{0.767} & \textcolor{gray}{5.507} \\
MDM 50steps  & Enc & & 0.455 & 0.645 & 0.749 & 0.489 & 3.330 \\
MDM & Enc & & 0.471 & 0.661 & 0.760 & 0.325 & 3.249 \\
MDM 50steps & Enc & $\checkmark$ & 0.489 & 0.685 & 0.787 & 0.355 & 3.100 \\
MDM & Enc & $\checkmark$& 0.463 & 0.658 & 0.705 & 0.265 & 3.266 \\
MDM 50steps & Dec & $\checkmark$ & \textbf{0.509} & \textbf{0.698} & \textbf{0.793} & 0.230 & 3.035 \\
MDM & Dec & $\checkmark$ & 0.502 & 0.694 & \textbf{0.793} & \textbf{0.165} & \textbf{3.020} \\
\midrule
% T2MGPT (Paper) & AR & & 0.491 & 0.680 & 0.775 & 0.116 & 3.118 \\
T2MGPT $\tau=0$  & AR & & 0.417 & 0.589 & 0.685 & 0.140 & 3.730 \\
T2MGPT $\tau=0.5$  & AR & & 0.491 & 0.680 & 0.775 & 0.116 & 3.118 \\
T2MGPT $\tau=[0, 1]$  & AR & & 0.492 & 0.679 & 0.775 & 0.141 & 3.121 \\
% T2MGPT & AR & & 0.484 & 0.672 & 0.770 & 0.117 & 3.153  \\
T2MGPT $\tau=0$ & AR & $\checkmark$ & 0.465 & 0.651 & 0.746 & 0.117 & 3.308  \\
T2MGPT $\tau=0.5$ & AR & $\checkmark$ & \textbf{0.539} & \textbf{0.730} & \textbf{0.823} & \textbf{0.105} & \textbf{2.838}  \\
T2MGPT $\tau=[0,1]$ & AR & $\checkmark$ & 0.538 & 0.730 & 0.821 & \textbf{0.105} & 2.841  \\
\midrule
CoMo  & AR & & 0.502 & 0.692 & 0.790 & 0.262 & 3.032  \\
CoMo no keywords  & AR & & 0.487 & - & - & 0.263 & 3.044  \\
CoMo  & AR & $\checkmark$ & \textbf{0.545} & \textbf{0.741} & \textbf{0.835} & \textbf{0.200} & \textbf{2.747}  \\
CoMo no keywords  & AR & $\checkmark$ & 0.539 & 0.738 & 0.831 & 0.226 & 2.777  \\
\bottomrule
\label{tab:quant_res5}
\vspace{-10pt}
\end{tabular}
\end{table}

\begin{figure*}[!t]
    \centering
    \includegraphics[width=\textwidth]{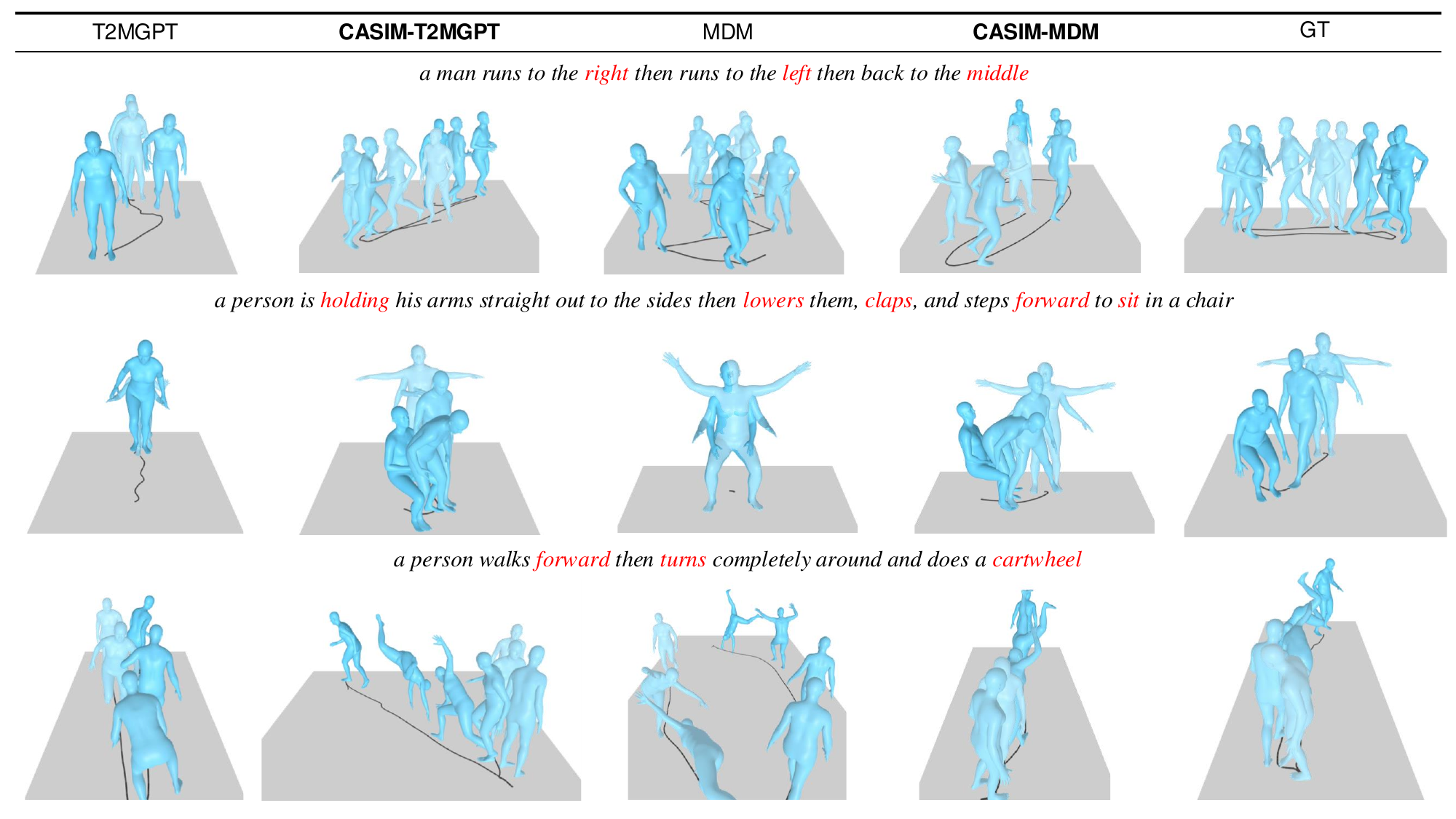}
    \vspace{-15pt}
    \captionof{figure}{Qualitative comparison between two baselines, their {\modulename}-enhanced models, and ground truth (GT) on HumanML3D test prompts. 
    Action verbs and their modifiers are highlighted in red, with motion sequences shown in color gradients (light to dark) and root trajectories in black.
    {\modulename}-MDM and {\modulename}-T2MGPT generate the motions that better match the descriptions, showing stronger text-motion correspondence and better controllability.}
    \vspace{-5pt}
\label{fig:qual_res}
\end{figure*}

\vspace{-3pt}
\subsection{Quantitative Analysis} 
\label{sub:quantitative}
\vspace{-3pt}

\textbf{Comparison with SOTA Methods.}
For quantitative evaluation, we report the results for each metric averaged over 20 repeated iterations on both datasets. 
Tab.~\ref{tab:quant_res1} and \ref{tab:quant_res2} present quantitative comparison of {\modulename}-MDM and {\modulename}-T2MGPT with the SOTA text-to-motion generation methods on both datasets.
Our method outperforms most SOTA methods in terms of R-Precision and MM-Dist, showing the effectiveness and robustness of {\modulename} in learning the text-motion correspondence.
In terms of motion quality, both {\modulename}-MDM and {\modulename}-T2MGPT achieve comparable FID score with some leading methods, such as GraphMotion, FineMoGen, and T2MGPT, on both benchmarks. 
Notably, the positive results are achieved through our semantic injection mechanism alone, without additional heuristic knowledge or textual semantics from external source like GraphMotion, CoMo, and FineMoGen.

\textbf{Leveling up SOTA model performances.}
To demonstrate {\modulename}'s effectiveness across different architectures, we integrate it with five representative models and evaluate their performance improvements. 
As shown in Tab.~\ref{tab:quant_res3}, {\modulename} consistently enhances all baseline models on HumanML3D dataset, with particularly notable gains in text-motion alignment metrics (R-Precision and MM-Dist).

For diffusion-based models, {\modulename} brings substantial improvements. MDM sees a significant boost in both motion quality (FID: 0.325→0.165) and multimodality alignment (Top1 R-Precision: 0.471→0.502, MM-Dist: 3.249→3.020). 
Similar improvements are observed for MotionDiffuse, suggesting that {\modulename}'s semantic injection mechanism effectively addresses the limitations of fixed-length text injection in diffusion models.
Autoregressive models, despite their stronger baseline performance, also benefit from {\modulename}. 
T2MGPT and CoMo show remarkable improvements in text-motion matching, and seems to benefit more with the composite aware semantics. 
The performance in Top1 R-Precision increases by 5.5\% and 4.3\% respectively and their MM-Distance drops to the lowest among all methods, with 2.838 and 2.747 respectively.
Even MoMask, which already achieves strong FID scores, sees consistent improvements across all metrics.
The benefits of {\modulename} remain evident on the more challenging KIT-ML dataset (Tab.~\ref{tab:quant_res4}). 
{\modulename}-MDM achieves significant improvements in text-motion alignment (R-Precision Top1: 0.164→0.448 and Top3: 0.396→0.784) and motion quality (FID: 0.497→0.354). 
{\modulename}-T2MGPT also sees consistent performance gain across the metrics.
For {\modulename}-CoMo,
% \footnote{CoMo's extra fine-grained keyword not open-sourced for KIT-ML dataset}
the motion quality remains on par with the baseline while R-Precision and MM-Distance improves.

The quantitative results suggest that {\modulename} is effective across different genres of motion generation models and across different datasets.
The consistent improvement in text-motion matching demonstrates its capability in learning dynamic semantic relationships without compromising motion quality.

\textbf{Architecture and configuration analysis.}
We conduct extensive experiments to validate {\modulename}'s effectiveness across different architectural choices and configuration settings, as shown in Tab.~\ref{tab:quant_res5}.
For diffusion-based MDM, we explore both encoder and decoder-based implementations of {\modulename}. Both variants demonstrate substantial improvements over their baseline (Encoder: Top1 R-Precision 0.471→0.489, FID 0.325→0.265; Decoder: Top3 R-Precision 0.608→0.793, FID 0.767→0.165), showcasing {\modulename}'s adaptability to different architectural choices. Particularly noteworthy is that these improvements hold even with significantly reduced computation--using only 50 diffusion steps instead of 1000, both variants maintain strong performance gains while achieving 20× faster inference.

For autoregressive models like T2MGPT, we examine {\modulename}'s behavior under different teacher forcing settings. During training, random masking ($\tau=[0,1]$ or $\tau=0.5$) \footnote{$\tau$ variable here represent coefficient for teacher forcing, which follows original notation. This is different from diffusion step index used earlier.} helps bridge the gap between training and inference, where the predicted tokens may differ from the ground truth. 
While the best performance occurs when $\tau=0.5$, {\modulename} can further improve its text-motion alignment (R-Precision: 0.491→0.539, MM-Dist: 3.118→2.838).
{\modulename} demonstrates robust performance across these hyperparameter configurations,
with all $\tau=0$, $\tau=0.5$ and $\tau = [0,1]$ 
achieving significant improvements, showing its resilience to different autoregressive model configurations.

The CoMo experiments reveal another interesting aspect of {\modulename}'s capabilities. 
The original CoMo relies on 11 additional keywords per description, augmented through GPT-4 to provide detailed motion characteristics across body parts and styles. 
Remarkably, {\modulename}-CoMo without any keywords outperforms the keyword-augmented baseline (R-Precision: 0.539 vs 0.487, FID: 0.226 vs 0.263), demonstrating {\modulename}'s ability to extract rich motion semantics directly from the input text without requiring external semantic augmentation.

Across all experiments, {\modulename} shows consistent performance improvements regardless of model architecture (diffusion or autoregressive), configuration choices (encoder/decoder, teacher forcing rates), or additional inputs (with/without keywords). This robust adaptability suggests that {\modulename}'s semantic injection mechanism provides fundamental improvements in learning text-motion correspondence that generalize across different modeling approaches.

\begin{figure}[!t]
    \centering
    \includegraphics[width=\linewidth]{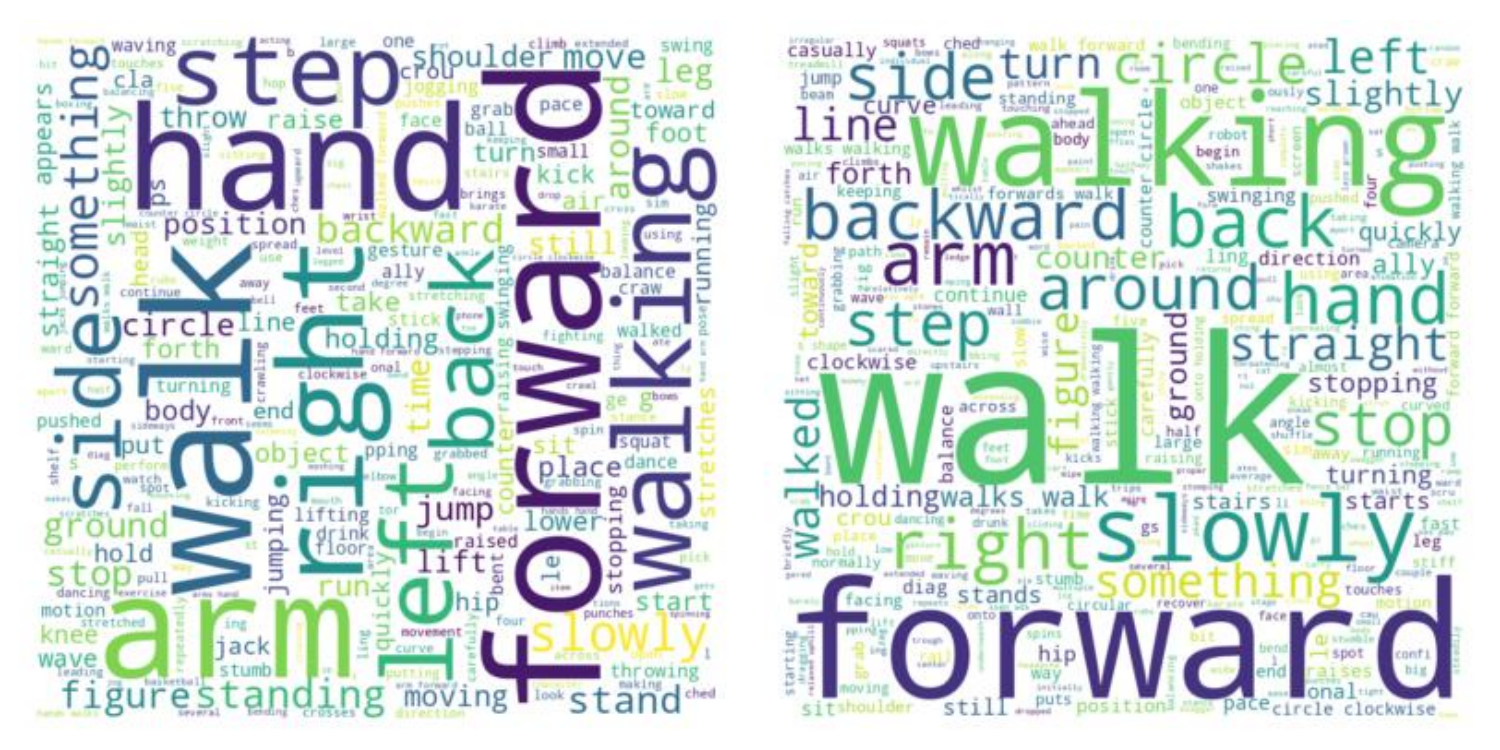}
    \vspace{-20pt}
    \captionof{figure}{Analysis of attention patterns in {\modulename}. 
    Left: Word cloud showing top-5 attended words across all test prompts, highlighting focus on action verbs, motion modifiers, and spatial references. 
    Right: Word cloud for prompts containing `\textit{walk}', revealing attention to motion-specific contextual attributes.}
    \vspace{-10pt}
\label{fig:wordcloud}
\end{figure}

\vspace{-3pt}
\subsection{Qualitative Analysis}
\label{sub:qualitative}
\vspace{-1pt}

\begin{table*}[t!]
\fontsize{8.25pt}{8.25pt}\selectfont
  \aboverulesep=0ex
  \belowrulesep=0.5ex 
\setlength{\tabcolsep}{6.0pt}
\centering
\caption{Results on the HumanML3D dataset for long-term motion generation.}
 \vspace{-5pt}
\begin{tabular}{lcccccccc|cc}
\toprule
\multirow{2}{*}{Method} & \multirow{2}{*}{\makecell{Handshake\\(\#frames)}} & \multirow{2}{*}{\modulename} & \multicolumn{6}{c}{\textbf{Motion}} & \multicolumn{2}{c}{\textbf{Transition}} \\
\cmidrule{4-11}
 & & & Top1$\uparrow$ & Top2$\uparrow$ & Top3$\uparrow$ & FID$\downarrow$ & MM-Dist$\downarrow$ & Div.$\rightarrow$& FID$\downarrow$ & Div.$\rightarrow$ \\
\midrule
GT & & & 0.511 & 0.703 & 0.797 & 0.002 & 2.974 & 9.503 & 0.050 & 9.570 \\
 \cmidrule{1-11}
\multirow{6}{*}{DoubleTake} & \multirow{2}{*}{20} &  & 0.309 & 0.477 & 0.589 & 0.953 & 5.713 & 9.624 & \textbf{1.540} & 8.750 \\
 &  & $\checkmark$ & \textbf{0.358} & \textbf{0.526} & \textbf{0.627} & \textbf{0.463} & \textbf{5.508} & \textbf{9.668} & 2.052 & \textbf{8.775} \\
 \cmidrule{2-11}
 & \multirow{2}{*}{30} & & -- & -- & 0.600 & 1.030 & 5.600 & 9.530 & 2.220 & 8.640 \\
 &  & $\checkmark$  & \textbf{0.345} & \textbf{0.516} & \textbf{0.612} & \textbf{0.707} & \textbf{5.532} & \textbf{9.864} & \textbf{1.896} & \textbf{8.805} \\
 \cmidrule{2-11}
 & \multirow{2}{*}{40} & & -- & -- & 0.580 & 1.160 & 5.670 & 9.610 & 2.410 & 8.610 \\
 &  & $\checkmark$ & \textbf{0.330} & \textbf{0.497} & \textbf{0.594} & \textbf{0.959} & \textbf{5.584} & \textbf{9.897} & \textbf{1.905} & \textbf{8.723} \\
\bottomrule
\end{tabular}%
\label{tab:dt_res}
 \vspace{-5pt}
\end{table*}

\textbf{Qualitative Comparisons.}
Fig.~\ref{fig:qual_res} demonstrates {\modulename}'s ability to improve motion generation quality through representative examples from the HumanML3D test set. Compared to baseline models, both {\modulename}-MDM and {\modulename}-T2MGPT show superior ability in following complex action sequences and maintaining temporal order. 
Their generated motions closely align with ground truth (GT), particularly in capturing the nuaunced semantics for spatial relationships and action transitions.

Specifically, for the input text ``\textit{a man runs to the right then runs to the left then back to the middle}", both {\modulename}-MDM and {\modulename}-T2MGPT accurately capture directional changes and chronological order, while baseline models struggle with spatial positioning and temporal progression. 
For another text, ``\textit{a person is holding his arms straight out to the sides then lowers them, claps, and steps forward to sit in a chair}", our models precisely follow each action component in sequence, whereas baseline models either miss critical components or fail to maintain the correct order. 
Given the third prompt, ``\textit{a person walks forward then turns completely around and does a cartwheel}", {\modulename}-MDM and {\modulename}-T2MGPT successfully reproduce the complete action sequence including the cartwheel motion, despite its infrequent appearance in the dataset.

The qualitative results demonstrate {\modulename}'s effectiveness in enabling precise text-based motion control while maintaining generalization to less common actions. 
The improved text-motion alignment across various examples suggests that our dynamic semantic injection successfully addresses the limitations of fixed-length text representations in existing methods.

\textbf{Interpretability Analysis.}
With the impressive quantitative and qualitative performance, it is interesting to understand how {\modulename} processes textual information. 
We first analyze its attention patterns through word clouds. 
Fig.~\ref{fig:wordcloud} (left) visualizes the most attended words in HumanML3D test set descriptions. The left word cloud shows the top-five attended words across all text prompts, revealing {\modulename}'s focus on motion-critical elements: action verbs (e.g., ``hand", ``step"), motion modifiers (e.g., ``slowly", ``quickly"), and spatial references (e.g., ``forward", ``circle"). For text containing the word `\textit{walk}' (Fig.~\ref{fig:wordcloud} right), the attention focuses on contextual motion attributes like direction (``forward", ``backward") and style (``slowly"), demonstrating {\modulename}'s capability in capturing motion-specific semantic relationships.

We also visualize the attention weights to analyze how {\modulename} dynamically aligns text with motion frames (Appendix \ref{sup:vis_attn_weights}). 
The visualization reveals clear temporal correspondence between text tokens and motion progression, validating our design of composite aware semantic injection in learning complex and dynamic text-motion relationships.

\vspace{-3pt}
% \subsection{Long-term Motion Generation}
\subsection{Long-term Motion Generation}
\label{sub:long-term}
\vspace{-1pt}

% While {\modulename} primarily targets single motion generation, we explore its effectiveness in long-term motion generation using DoubleTake \cite{shafir2024human}, a framework that employs a two-stage diffusion process to generate smooth transitions between MDM-generated motions.
% As shown in Tab. \ref{tab:dt_res}, {\modulename} enhances motion quality, text-motion matching, and diversity across different handshake sizes in the generated motion clips, demonstrating its benefits beyond single motion generation.
% However, when evaluated on transition periods between motion clips, the improvements are less consistent and vary with handshake size.
% This mixed performance on transitions, while not directly addressed by {\modulename}'s design, opens interesting questions about how semantic injection methods influence diffusion-based transition generation in long-term motion synthesis.

While {\modulename} primarily targets single motion generation, we explore its effectiveness in long-term motion generation using DoubleTake \cite{shafir2024human}, a framework that employs an additional two-stage diffusion process to generate smooth transitions between the motion clips generated by diffusion models like MDM.

As shown in Tab.~\ref{tab:dt_res}, {\modulename} consistently improves the motion clips across metrics and handshake sizes overall. 
For handshake size of 20 frames, {\modulename} significantly reduces the motion FID from 0.953 to 0.463 while improving text alignment (Top1: 0.309→0.358) and maintaining motion diversity (9.624→9.668). 
Similar improvements are observed with longer handshake periods of 30 and 40 frames, though the gains in FID gradually decrease as the transition period extends.
Interestingly, the transition quality shows mixed results. With 20-frame handshake, the transition FID slightly increases (1.540→2.052), while longer handshakes see improvements (30 frames: 2.220→1.896; 40 frames: 2.410→1.905). 
This suggests that {\modulename}'s semantic injection, while effective for individual motion generation, interacts differently for compositing several motion clips during the motion blending and transition generation process.
The observations raise interesting questions about how semantic injection methods influence diffusion-based transition generation in long-term motion synthesis, particularly regarding the balance between local motion quality and smooth transitions.

%%% 1 or 2 figures with qualitative samples and comparison for different methods, in-domain, spatial control, long text.

\vspace{-5pt}
\section{Conclusion} \label{sec:conclusion}
\vspace{-3pt}

% We propose {\modulename}, a simple yet effective composite aware semantic injection method, that is agnostic to various genres of text-to-motion generation models and motion representations, spanning across autoregressive to denoising diffusion and from descritized motion token to noisy motion sequence.
We propose {\modulename}, a simple yet effective method for semantic injection that works with various text-to-motion models and representations, from autoregressive to denoising diffusion and discretized motion tokens to continuous raw motion sequences.
Our experiments suggest that {\modulename} consistently improves the motion quality and strengthens the text-motion alignment across several state-of-the-art models on HumanML3D and KIT benchmarks.
The method shows promise in enhancing long-term human motion generation. 
Notably, it enables more precise motion control through input text compared to fixed-length semantic injection approaches.

\textbf{Limitations and Future Work.} While our semantic injection method shows potential for processing very long text inputs for zero-shot long-term motion generation, it sill relies on motion blending techniques such as DoubleTake.
% it does not significantly improve the performance without additional motion blending methods such as DoubleTake.
This limitation largely arises from the training dataset itself, which lacks long text and motion samples. 
Future work would focus on curating datasets with extended text-motion pairs and developing techniques to effectively leverage such data.
Though {\modulename} preserves the composite nature for text injection, it shows limited improvements with methods like MLD~\cite{chen2023executing} that encode motion as a fixed-length latent vector.
This compression itself constrains the learning of fine-grained text-motion correspondence. Please refer to Supp. Mat. for more discussion.

\vspace{-5pt}
\section*{Impact Statement} \label{sec:impact_statement}
\vspace{-3pt}

Our work on text-to-motion generation aims to advance machine learning and its relevant applications. 
The primary positive impact lies in enabling more intuitive and accessible ways to create human animations, potentially benefiting creative industries, interactive content creation, and assistive technologies. 
However, we acknowledge potential risks with the generation of inappropriate or misleading motions if the system is misused. 
While our method focuses on improving technical capabilities rather than introducing new application scenarios, we encourage future research and development to implement appropriate content filtering and user verification mechanisms to ensure a safe and responsible use of this technology.

% Authors are \textbf{required} to include a statement of the potential 
% broader impact of their work, including its ethical aspects and future 
% societal consequences. This statement should be in an unnumbered 
% section at the end of the paper (co-located with Acknowledgements -- 
% the two may appear in either order, but both must be before References), 
% and does not count toward the paper page limit. In many cases, where 
% the ethical impacts and expected societal implications are those that 
% are well established when advancing the field of Machine Learning, 
% substantial discussion is not required, and a simple statement such 
% as the following will suffice:

% ``This paper presents work whose goal is to advance the field of 
% Machine Learning. There are many potential societal consequences 
% of our work, none which we feel must be specifically highlighted here.''

% The above statement can be used verbatim in such cases, but we 
% encourage authors to think about whether there is content which does 
% warrant further discussion, as this statement will be apparent if the 
% paper is later flagged for ethics review.

% Impact Statement section does not count towards the 8-page limit
%%% Acknowledgement is not needed till camera ready version
\vspace{-5pt}
\section*{Acknowledgement} \label{sec:acknowledgment}
\vspace{-3pt}
The research was supported in part by NSF awards: IIS-1703883, IIS-1955404, IIS-1955365, RETTL-2119265, and EAGER-2122119. This work was also partially supported by the Center for Smart Streetscapes, an NSF Engineering Research Center, under cooperative agreement EEC-2133516.
This material is based upon work supported by the U.S. Department of Homeland Security\footnote{Disclaimer. The views and conclusions contained in this document are those of the authors and should not be interpreted as necessarily representing the official policies, either expressed or implied, of the U.S. Department of Homeland Security.} under Grant Award Number 22STESE00001 01 01.

\nocite{langley00}

\bibliography{reference}

\begin{thebibliography}{39}
\providecommand{\natexlab}[1]{#1}
\providecommand{\url}[1]{\texttt{#1}}
\expandafter\ifx\csname urlstyle\endcsname\relax
  \providecommand{\doi}[1]{doi: #1}\else
  \providecommand{\doi}{doi: \begingroup \urlstyle{rm}\Url}\fi

\bibitem[Alexanderson et~al.(2023)Alexanderson, Nagy, Beskow, and Henter]{alexanderson2023listen}
Alexanderson, S., Nagy, R., Beskow, J., and Henter, G.~E.
\newblock Listen, denoise, action! audio-driven motion synthesis with diffusion models.
\newblock \emph{ACM Trans. Graph.}, 42\penalty0 (4):\penalty0 44:1--44:20, 2023.
\newblock \doi{10.1145/3592458}.

\bibitem[Athanasiou et~al.(2024)Athanasiou, Cseke, Diomataris, Black, and Varol]{MotionFix}
Athanasiou, N., Cseke, A., Diomataris, M., Black, M.~J., and Varol, G.
\newblock Motionfix: Text-driven 3d human motion editing.
\newblock In \emph{SIGGRAPH Asia 2024 Conference Proceedings}. ACM, December 2024.
\newblock URL \url{https://motionfix.is.tue.mpg.de/}.

\bibitem[Cen et~al.(2024)Cen, Pi, Peng, Shen, Yang, Shuai, Bao, and Zhou]{cen2024text_scene_motion}
Cen, Z., Pi, H., Peng, S., Shen, Z., Yang, M., Shuai, Z., Bao, H., and Zhou, X.
\newblock Generating human motion in 3d scenes from text descriptions.
\newblock In \emph{CVPR}, 2024.

\bibitem[Chang et~al.(2022)Chang, Zhang, and Kapadia]{chang2022ivi}
Chang, C.-J., Zhang, S., and Kapadia, M.
\newblock The ivi lab entry to the genea challenge 2022--a tacotron2 based method for co-speech gesture generation with locality-constraint attention mechanism.
\newblock In \emph{Proceedings of the 2022 International Conference on Multimodal Interaction}, pp.\  784--789, 2022.

\bibitem[Chang et~al.(2023)Chang, Sohn, Zhang, Jayashankar, Usman, and Kapadia]{chang2023importance}
Chang, C.-J., Sohn, S.~S., Zhang, S., Jayashankar, R., Usman, M., and Kapadia, M.
\newblock The importance of multimodal emotion conditioning and affect consistency for embodied conversational agents.
\newblock In \emph{Proceedings of the 28th International Conference on Intelligent User Interfaces}, pp.\  790--801, 2023.

\bibitem[Chang et~al.(2024{\natexlab{a}})Chang, Li, Moon, and Kapadia]{chang2024equivalency}
Chang, C.-J., Li, D., Moon, S., and Kapadia, M.
\newblock On the equivalency, substitutability, and flexibility of synthetic data.
\newblock \emph{arXiv preprint arXiv:2403.16244}, 2024{\natexlab{a}}.

\bibitem[Chang et~al.(2024{\natexlab{b}})Chang, Li, Patel, Goel, Zhou, Moon, Sohn, Yoon, Pavlovic, and Kapadia]{chang2024learning}
Chang, C.-J., Li, D., Patel, D., Goel, P., Zhou, H., Moon, S., Sohn, S.~S., Yoon, S., Pavlovic, V., and Kapadia, M.
\newblock Learning from synthetic human group activities.
\newblock In \emph{Proceedings of the IEEE/CVF Conference on Computer Vision and Pattern Recognition}, pp.\  21922--21932, 2024{\natexlab{b}}.

\bibitem[Chen et~al.(2024)Chen, Lu, Zeng, Zhang, Wang, Zhang, and Zhang]{chen2024motionllm}
Chen, L.-H., Lu, S., Zeng, A., Zhang, H., Wang, B., Zhang, R., and Zhang, L.
\newblock Motionllm: Understanding human behaviors from human motions and videos.
\newblock \emph{arXiv preprint arXiv:2405.20340}, 2024.

\bibitem[Chen et~al.(2023)Chen, Jiang, Liu, Huang, Fu, Chen, and Yu]{chen2023executing}
Chen, X., Jiang, B., Liu, W., Huang, Z., Fu, B., Chen, T., and Yu, G.
\newblock Executing your commands via motion diffusion in latent space.
\newblock In \emph{Proceedings of the IEEE/CVF Conference on Computer Vision and Pattern Recognition}, pp.\  18000--18010, 2023.

\bibitem[Delmas et~al.(2022)Delmas, Weinzaepfel, Lucas, Moreno-Noguer, and Rogez]{delmas2022posescript}
Delmas, G., Weinzaepfel, P., Lucas, T., Moreno-Noguer, F., and Rogez, G.
\newblock {PoseScript: 3D Human Poses from Natural Language}.
\newblock In \emph{{ECCV}}, 2022.

\bibitem[Devlin(2018)]{devlin2018bert}
Devlin, J.
\newblock Bert: Pre-training of deep bidirectional transformers for language understanding.
\newblock \emph{arXiv preprint arXiv:1810.04805}, 2018.

\bibitem[Dhariwal \& Nichol(2021)Dhariwal and Nichol]{dhariwal2021diffusion}
Dhariwal, P. and Nichol, A.
\newblock Diffusion models beat gans on image synthesis.
\newblock \emph{Advances in neural information processing systems}, 34:\penalty0 8780--8794, 2021.

\bibitem[Goel et~al.(2024)Goel, Wang, Liu, and Fatahalian]{goel2024iterative}
Goel, P., Wang, K.-C., Liu, C.~K., and Fatahalian, K.
\newblock Iterative motion editing with natural language.
\newblock In \emph{ACM SIGGRAPH 2024 Conference Papers}, pp.\  1--9, 2024.

\bibitem[Guo et~al.(2020)Guo, Zuo, Wang, Zou, Sun, Deng, Gong, and Cheng]{guo2020action2motion}
Guo, C., Zuo, X., Wang, S., Zou, S., Sun, Q., Deng, A., Gong, M., and Cheng, L.
\newblock Action2motion: Conditioned generation of 3d human motions.
\newblock In \emph{Proceedings of the 28th ACM International Conference on Multimedia}, pp.\  2021--2029, 2020.

\bibitem[Guo et~al.(2022{\natexlab{a}})Guo, Zou, Zuo, Wang, Ji, Li, and Cheng]{Guo_2022_CVPR}
Guo, C., Zou, S., Zuo, X., Wang, S., Ji, W., Li, X., and Cheng, L.
\newblock Generating diverse and natural 3d human motions from text.
\newblock In \emph{Proceedings of the IEEE/CVF Conference on Computer Vision and Pattern Recognition (CVPR)}, pp.\  5152--5161, June 2022{\natexlab{a}}.

\bibitem[Guo et~al.(2022{\natexlab{b}})Guo, Zuo, Wang, and Cheng]{guo2022tm2t}
Guo, C., Zuo, X., Wang, S., and Cheng, L.
\newblock Tm2t: Stochastic and tokenized modeling for the reciprocal generation of 3d human motions and texts.
\newblock In \emph{European Conference on Computer Vision}, pp.\  580--597. Springer, 2022{\natexlab{b}}.

\bibitem[Guo et~al.(2023)Guo, Mu, Javed, Wang, and Cheng]{guo2023momask}
Guo, C., Mu, Y., Javed, M.~G., Wang, S., and Cheng, L.
\newblock Momask: Generative masked modeling of 3d human motions.
\newblock 2023.

\bibitem[Huang et~al.(2024)Huang, Wan, Yang, Callison-Burch, Yatskar, and Liu]{huang2024como}
Huang, Y., Wan, W., Yang, Y., Callison-Burch, C., Yatskar, M., and Liu, L.
\newblock Como: Controllable motion generation through language guided pose code editing, 2024.

\bibitem[Jiang et~al.(2024)Jiang, Chen, Liu, Yu, Yu, and Chen]{jiang2024motiongpt}
Jiang, B., Chen, X., Liu, W., Yu, J., Yu, G., and Chen, T.
\newblock Motiongpt: Human motion as a foreign language.
\newblock \emph{Advances in Neural Information Processing Systems}, 36, 2024.

\bibitem[Jin et~al.(2023)Jin, Wu, Fan, Sun, Wei, and Yuan]{jin2023act}
Jin, P., Wu, Y., Fan, Y., Sun, Z., Wei, Y., and Yuan, L.
\newblock Act as you wish: Fine-grained control of motion diffusion model with hierarchical semantic graphs.
\newblock In \emph{NeurIPS}, 2023.

\bibitem[Karunratanakul et~al.(2023)Karunratanakul, Preechakul, Suwajanakorn, and Tang]{karunratanakul2023gmd}
Karunratanakul, K., Preechakul, K., Suwajanakorn, S., and Tang, S.
\newblock Guided motion diffusion for controllable human motion synthesis.
\newblock In \emph{Proceedings of the IEEE/CVF International Conference on Computer Vision}, pp.\  2151--2162, 2023.

\bibitem[Lee et~al.(2024)Lee, Baradel, Lucas, Lee, and Rogez]{lee2024t2lm}
Lee, T., Baradel, F., Lucas, T., Lee, K.~M., and Rogez, G.
\newblock T2lm: Long-term 3d human motion generation from multiple sentences.
\newblock In \emph{Proceedings of the IEEE/CVF Conference on Computer Vision and Pattern Recognition}, pp.\  1867--1876, 2024.

\bibitem[Li et~al.(2024)Li, Chibane, He, Pearl, Geiger, and Pons-Moll]{li2024unimotion}
Li, C., Chibane, J., He, Y., Pearl, N., Geiger, A., and Pons-Moll, G.
\newblock Unimotion: Unifying 3d human motion synthesis and understanding.
\newblock \emph{arXiv preprint arXiv:2409.15904}, 2024.

\bibitem[Mahmood et~al.(2019)Mahmood, Ghorbani, Troje, Pons-Moll, and Black]{mahmood2019amass}
Mahmood, N., Ghorbani, N., Troje, N.~F., Pons-Moll, G., and Black, M.~J.
\newblock Amass: Archive of motion capture as surface shapes.
\newblock In \emph{Proceedings of the IEEE/CVF international conference on computer vision}, pp.\  5442--5451, 2019.

\bibitem[OpenAI et~al.(2024)OpenAI, Achiam, Adler, et~al.]{openai2024gpt4technicalreport}
OpenAI, Achiam, J., Adler, S., et~al.
\newblock Gpt-4 technical report, 2024.
\newblock URL \url{https://arxiv.org/abs/2303.08774}.

\bibitem[Plappert et~al.(2016)Plappert, Mandery, and Asfour]{Plappert2016}
Plappert, M., Mandery, C., and Asfour, T.
\newblock The {KIT} motion-language dataset.
\newblock \emph{Big Data}, 4\penalty0 (4):\penalty0 236--252, dec 2016.
\newblock \doi{10.1089/big.2016.0028}.
\newblock URL \url{http://dx.doi.org/10.1089/big.2016.0028}.

\bibitem[Radford et~al.(2018)Radford, Narasimhan, Salimans, and Sutskever]{radford2018improving}
Radford, A., Narasimhan, K., Salimans, T., and Sutskever, I.
\newblock Improving language understanding by generative pre-training.
\newblock 2018.

\bibitem[Radford et~al.(2021)Radford, Kim, Hallacy, Ramesh, Goh, Agarwal, Sastry, Askell, Mishkin, Clark, et~al.]{radford2021learning}
Radford, A., Kim, J.~W., Hallacy, C., Ramesh, A., Goh, G., Agarwal, S., Sastry, G., Askell, A., Mishkin, P., Clark, J., et~al.
\newblock Learning transferable visual models from natural language supervision.
\newblock In \emph{International conference on machine learning}, pp.\  8748--8763. PMLR, 2021.

\bibitem[Setareh et~al.(2024)Setareh, Tevet, Reda, Peng, and van~de Panne]{CondMDICohan2024}
Setareh, C., Tevet, G., Reda, D., Peng, X.~B., and van~de Panne, M.
\newblock Generating human interaction motions in scenes with text control.
\newblock 2024.

\bibitem[Shafir et~al.(2024)Shafir, Tevet, Kapon, and Bermano]{shafir2024human}
Shafir, Y., Tevet, G., Kapon, R., and Bermano, A.~H.
\newblock Human motion diffusion as a generative prior.
\newblock In \emph{The Twelfth International Conference on Learning Representations}, 2024.

\bibitem[Shi et~al.(2023)Shi, Luo, Peng, Zhang, and Sun]{shi2023generating}
Shi, X., Luo, C., Peng, J., Zhang, H., and Sun, Y.
\newblock Generating fine-grained human motions using chatgpt-refined descriptions.
\newblock \emph{arXiv preprint arXiv:2312.02772}, 2023.

\bibitem[Stathopoulos et~al.(2024)Stathopoulos, Han, and Metaxas]{stathopoulos2024score}
Stathopoulos, A., Han, L., and Metaxas, D.
\newblock Score-guided diffusion for 3d human recovery.
\newblock In \emph{CVPR}, 2024.

\bibitem[Tevet et~al.(2023)Tevet, Raab, Gordon, Shafir, Cohen-or, and Bermano]{tevet2023human}
Tevet, G., Raab, S., Gordon, B., Shafir, Y., Cohen-or, D., and Bermano, A.~H.
\newblock Human motion diffusion model.
\newblock In \emph{The Eleventh International Conference on Learning Representations}, 2023.
\newblock URL \url{https://openreview.net/forum?id=SJ1kSyO2jwu}.

\bibitem[Van Den~Oord et~al.(2017)Van Den~Oord, Vinyals, et~al.]{van2017neural}
Van Den~Oord, A., Vinyals, O., et~al.
\newblock Neural discrete representation learning.
\newblock \emph{Advances in neural information processing systems}, 30, 2017.

\bibitem[Zhang et~al.(2023{\natexlab{a}})Zhang, Zhang, Cun, Huang, Zhang, Zhao, Lu, and Shen]{zhang2023generating}
Zhang, J., Zhang, Y., Cun, X., Huang, S., Zhang, Y., Zhao, H., Lu, H., and Shen, X.
\newblock T2m-gpt: Generating human motion from textual descriptions with discrete representations.
\newblock In \emph{Proceedings of the IEEE/CVF Conference on Computer Vision and Pattern Recognition (CVPR)}, 2023{\natexlab{a}}.

\bibitem[Zhang et~al.(2022)Zhang, Cai, Pan, Hong, Guo, Yang, and Liu]{zhang2022motiondiffuse}
Zhang, M., Cai, Z., Pan, L., Hong, F., Guo, X., Yang, L., and Liu, Z.
\newblock Motiondiffuse: Text-driven human motion generation with diffusion model.
\newblock \emph{arXiv preprint arXiv:2208.15001}, 2022.

\bibitem[Zhang et~al.(2023{\natexlab{b}})Zhang, Li, Cai, Ren, Yang, and Liu]{zhang2023finemogen}
Zhang, M., Li, H., Cai, Z., Ren, J., Yang, L., and Liu, Z.
\newblock Finemogen: Fine-grained spatio-temporal motion generation and editing.
\newblock \emph{NeurIPS}, 2023{\natexlab{b}}.

\bibitem[Zhao et~al.(2019)Zhao, Peng, Tian, Kapadia, and Metaxas]{zhao2019semantic}
Zhao, L., Peng, X., Tian, Y., Kapadia, M., and Metaxas, D.~N.
\newblock Semantic graph convolutional networks for 3d human pose regression.
\newblock In \emph{Proceedings of the IEEE/CVF conference on computer vision and pattern recognition}, pp.\  3425--3435, 2019.

\bibitem[Zhou et~al.(2024)Zhou, Wan, and Wang]{zhou2024avatargpt}
Zhou, Z., Wan, Y., and Wang, B.
\newblock Avatargpt: All-in-one framework for motion understanding planning generation and beyond.
\newblock In \emph{Proceedings of the IEEE/CVF Conference on Computer Vision and Pattern Recognition}, pp.\  1357--1366, 2024.

\end{thebibliography}
\bibliographystyle{icml2025}

%%% APPENDIX
\appendix
\onecolumn

\section{Additional Qualitative Results} \label{sup:add_qual_res}
% Refer to the Demo Video
We provide a supplementary video on our project page, demonstrating side-by-side comparisons between baseline models (MDM, T2MGPT) and their {\modulename}-enhanced versions ({\modulename}-MDM, {\modulename}-T2MGPT). 
The video showcases the superiority of our method as 
{\modulename} enables more precise control over generated motions through text prompts, accurately capturing nuanced word differences and maintaining proper chronological order.

\section{Ablation Study}

\subsection{Different Text Encoder for {\modulename}}
To understand how different text encoders affect {\modulename}'s performance, we compare CLIP~\cite{radford2021learning} with BERT~\cite{devlin2018bert} using MDM as the base model. 
As shown in Tab.~\ref{tab:quant_bert}, BERT-based {\modulename} achieves better performance in text-motion alignment metrics (R-Precision: 0.511 vs. 0.478, MM-Dist.: 2.938 vs. 3.272) and motion diversity (9.630 vs. 9.468). 
While CLIP shows slightly better FID (0.303 vs. 0.346), the overall results suggest that BERT's contextualized word embeddings might be more suitable for capturing motion-relevant semantics.

This performance difference could be attributed to BERT's bidirectional context modeling and its pretraining objectives that focus on understanding relationships between word tokens, which aligns well with our goal of preserving composite-aware semantics and understanding the nuanced differences between token embeddings.

\begin{table}[h!]
\fontsize{9pt}{9pt}\selectfont
  \aboverulesep=0ex
  \belowrulesep=0.5ex 
\setlength{\tabcolsep}{8pt}
\centering
\caption{Quantitative results with different text encoder for {\modulename} on HumanML3D dataset.}
\begin{tabular}{lcccccccc}
\toprule
\multirow{2}{*}{Method} & \multirow{2}{*}{\modulename}  & \multirow{2}{*}{Text Encoder} & \multicolumn{3}{c}{R-Precision$\uparrow$} & \multirow{2}{*}{FID$\downarrow$} & \multirow{2}{*}{MM-Dist.$\downarrow$} & \multirow{2}{*}{Div.$\uparrow$} \\
 \cmidrule{4-6}
 & & & Top1 & Top2 & Top3 &  &   \\
\midrule
MDM & $\checkmark$ & BERT & \textbf{0.511} & \textbf{0.703} & \textbf{0.799} & 0.346 & \textbf{2.938} & \textbf{9.630} \\
MDM & $\checkmark$ & CLIP & 0.478 & 0.666 & 0.765 & \textbf{0.303} & 3.272 & 9.468 \\ 
\bottomrule
\label{tab:quant_bert}
\end{tabular}
\end{table}

\subsection{CLIP Embeddings from Different Layers for {\modulename}}

While our main experiments use CLIP's latent embeddings for semantic injection, it is interesting to understand how CLIP token embeddings from different layers for {\modulename} could potentially influence the performance of text-to-motion generation.
We compare the token embeddings from the output of the final layer and the output of the last layer before final projection.
Tab.~\ref{tab:quant_clip_latent} presents this comparison using MDM as the base model.
Using final layer embeddings shows stronger text-motion alignment (R-Precision Top1: 0.517 vs. 0.478) and lower multimodal distance (MM-Dist.: 2.945 vs. 3.272).
However, this comes at the cost of motion quality, as indicated by the higher FID score (0.410 vs. 0.303). 
This trade-off suggests that while the final-layer embeddings might better capture text-motion correspondences, the latent embeddings may provide a better balance between motion quality and semantic alignment.

\begin{table}[h!]
\fontsize{9pt}{9pt}\selectfont
  \aboverulesep=0ex
  \belowrulesep=0.5ex 
\setlength{\tabcolsep}{8pt}
\centering
\caption{Quantitative results with CLIP embeddings from different layers for {\modulename} on HumanML3D dataset.}
\begin{tabular}{lcccccccc}
\toprule
\multirow{2}{*}{Method} & \multirow{2}{*}{\modulename}  & \multirow{2}{*}{CLIP Embedding} & \multicolumn{3}{c}{R-Precision$\uparrow$} & \multirow{2}{*}{FID$\downarrow$} & \multirow{2}{*}{MM-Dist.$\downarrow$} & \multirow{2}{*}{Div.$\uparrow$} \\
 \cmidrule{4-6}
 & & & Top1 & Top2 & Top3 &  &   \\
\midrule
MDM & $\checkmark$ & Final & \textbf{0.517} & \textbf{0.709} & \textbf{0.806} & 0.410 & \textbf{2.945} & \textbf{9.735} \\
MDM & $\checkmark$ & Latent & 0.478 & 0.666 & 0.765 & \textbf{0.303} & 3.272 & 9.468 \\ 
\bottomrule
\label{tab:quant_clip_latent}
\end{tabular}
\end{table}

\section{Limitation with MLD} \label{sup:limit_mld}

We evaluate {\modulename} with MLD \cite{chen2023executing}, a diffusion-based method that operates in a learned latent space by compressing motion sequences into fixed-length vectors. 
As shown in Tab. \ref{tab:quant_mld}, {\modulename} provides limited improvements: while FID slightly improves (0.532→0.502), text-motion alignment metrics show marginal decreases (Top1 R-Precision: 0.469→0.452, MM-Distance: 3.282→3.389).
This performance pattern differs from our results with other models and can be attributed to MLD's choice of using fixed-length latent vectors for motion representation. 
While this design benefits motion editing tasks, it inherently limits the model's ability to capture composite motion structures. 
The observation suggests that {\modulename}'s effectiveness depends on the underlying motion representation's ability to preserve temporal and structural information.

\begin{table}[h!]
\fontsize{9pt}{9pt}\selectfont
  \aboverulesep=0ex
  \belowrulesep=0.5ex 
\setlength{\tabcolsep}{8pt}
\centering
\caption{Quantitative results with MLD and \modulename-MLD on HumanML3D dataset.}
\begin{tabular}{lccccccc}
\toprule
\multirow{2}{*}{Method} & \multirow{2}{*}{\modulename} & \multicolumn{3}{c}{R-Precision$\uparrow$} & \multirow{2}{*}{FID$\downarrow$} & \multirow{2}{*}{MM-Dist.$\downarrow$} & \multirow{2}{*}{Div.$\uparrow$} \\
 \cmidrule{3-5}
 & & Top1 & Top2 & Top3 &  &   \\
\midrule
MLD & & \textbf{0.469} & \textbf{0.659} & \textbf{0.760} & 0.532 & \textbf{3.282} & \textbf{9.570} \\
MLD & $\checkmark$ & 0.452 & 0.637 & 0.740 & \textbf{0.502} & 3.389 & 9.132 \\ 
\bottomrule
\label{tab:quant_mld}
\end{tabular}
\end{table}

\section{Visualization of Text-Motion Attention} \label{sup:vis_attn_weights}

% To understand how {\modulename} supports text-motion alignment, we analyze cross-attention in the \modulename-MDM model. 
% Figure \ref{fig:attn} shows the attention heatmap between generated motion and text at the first diffusion step. Early keyframes focus attention on initial words matching the waving motion. As keyframes progress, attention shifts to later text keywords, like `\textit{sit}' and `\textit{down}', illustrating how {\modulename} helps the model generate precise temporal motions.

\begin{figure*}[!h]
    \centering
    \includegraphics[width=\textwidth]{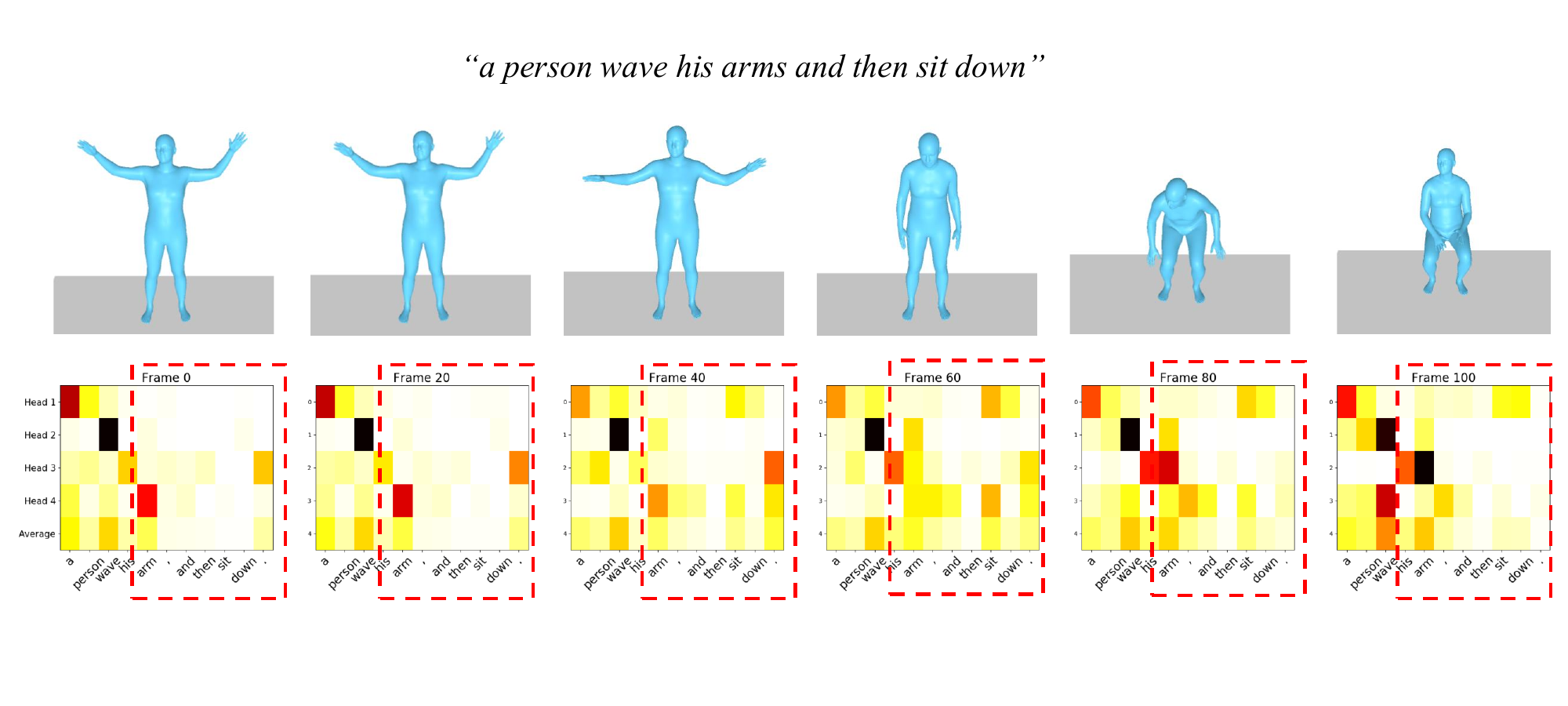}
    \vspace{-30pt}
    \caption{Visualization of attention weights in {\modulename}-MDM. 
    Top: Generated motion sequence for the prompt "\textit{a person wave his arms and then sit down}". 
    Bottom: Attention heatmaps for four attention heads and their average from the last layer. 
        % It shows the dynamic text-motion alignment as the sequence progresses from waving to sitting motion.
    }

    \vspace{-20pt}
\label{fig:attn}
\end{figure*}

We visualize the text-motion attention in {\modulename}-MDM to understand how text tokens influence motion generation at different motion frames. 
Figure \ref{fig:attn} shows the attention patterns for the prompt "\textit{a person wave his arms and then sit down}". 
The visualization demonstrates {\modulename}'s ability to capture temporal dependencies: early frames (0-40) attend strongly to words related to the waving motion, while later frames (60-100) shift attention to sit-related tokens. 
This progressive attention transition validates our design of composite-aware semantic injection for handling sequential motion descriptions.

%%% Visualization of attention weights 

% \section{User Study} \label{sub:user_study}

% \section{Additional Quantitative Evaluation} \label{sup:motioncritic}

\end{document}